\documentclass[10pt,twocolumn,letterpaper]{article}

\usepackage[pagenumbers]{cvpr}      

\usepackage[dvipsnames]{xcolor}
\usepackage{times}
\usepackage{epsfig}
\usepackage{graphicx}
\usepackage{amsmath}
\usepackage{amssymb}

\usepackage{algorithm}
\usepackage{algorithmic}

\usepackage{booktabs}       
\usepackage{amsfonts}       
\usepackage{nicefrac}       
\usepackage{microtype}      
\usepackage{comment}
\usepackage{tikz, siunitx}  
\usepackage{amsmath}    
\usepackage{amssymb}    
\usepackage{xspace}
\usepackage{tabularx}
\usepackage{threeparttable} 
\usepackage{rotating}
\usepackage{multirow}
\usepackage{subcaption}
\usepackage{makecell}
\usepackage[export]{adjustbox}
\usepackage{enumitem} 

\newcommand{\figlabel}{Fig.\xspace}

\newcommand{\minorsection}[1]{\noindent\textbf{#1.}}

\usepackage{pifont}
\definecolor{Highlight}{HTML}{39b54a}  
\newcommand{\hl}[1]{\textcolor{Highlight}{#1}}

\newcommand{\reshl}[3]{
{#1} \fontsize{0.8em}{0.8em}\selectfont{\hl{(${#2}$\textbf{#3})}}}

\usepackage{amssymb}
\usepackage{pifont}
\usepackage{xcolor, colortbl} 
\newcommand{\gr}{\rowcolor[gray]{.95}}
\newcommand{\grayrow}{\rowcolor[gray]{.95}}

\usepackage{algorithm}
\usepackage{listings}
\usepackage{etoolbox}
\makeatletter
\AfterEndEnvironment{algorithm}{\let\@algcomment\relax}
\AtEndEnvironment{algorithm}{\kern2pt\hrule\relax\vskip3pt\@algcomment}
\let\@algcomment\relax
\newcommand\algcomment[1]{\def\@algcomment{\footnotesize#1}}
\renewcommand\fs@ruled{\def\@fs@cfont{\bfseries}\let\@fs@capt\floatc@ruled
  \def\@fs@pre{\hrule height.8pt depth0pt \kern2pt}%
  \def\@fs@post{}%
  \def\@fs@mid{\kern2pt\hrule\kern2pt}%
  \let\@fs@iftopcapt\iftrue}
\makeatother
\newcommand{\cmmnt}[1]{}

\usepackage[normalem]{ulem}

\DeclareMathOperator*{\argmin}{arg\,min}

\newcommand{\sota}{state-of-the-art\xspace}

\definecolor{cvprblue}{rgb}{0.21,0.49,0.74}
\usepackage[pagebackref,breaklinks,colorlinks,citecolor=cvprblue]{hyperref}
\usepackage[capitalize]{cleveref}
\crefname{section}{Sec.}{Secs.}
\Crefname{section}{Section}{Sections}
\Crefname{table}{Table}{Tables}
\crefname{table}{Tab.}{Tabs.}

\def\paperID{62} 
\def\confName{3DV\xspace}
\def\confYear{2024\xspace}

\begin{document}

\title{Pix4Point: Image Pretrained Standard Transformers for 3D Point Cloud Understanding}

\author{
Guocheng Qian\thanks{\footnotesize{Equal contribution.}}~, Abdullah Hamdi$^*$, Xingdi Zhang$^*$, Bernard Ghanem\\
King Abdullah University of Science and Technology (KAUST)\\
\tt\small \{guocheng.qian,abdullah.hamdi,xingdi.zhang,bernard.ghanem\}@kaust.edu.sa}

\maketitle
\begin{abstract}
While Transformers have achieved impressive success in natural language processing and computer vision, their performance on 3D point clouds is relatively poor. This is mainly due to the limitation of Transformers: a demanding need for extensive training data. Unfortunately, in the realm of 3D point clouds, the availability of large datasets is a challenge, exacerbating the issue of training Transformers for 3D tasks. In this work, we solve the data issue of point cloud Transformers from two perspectives: (i) introducing more inductive bias to reduce the dependency of Transformers on data, and (ii) relying on cross-modality pretraining.  More specifically, we first present Progressive Point Patch Embedding and present a new point cloud Transformer model namely PViT. PViT shares the same backbone as Transformer but is shown to be less hungry for data, enabling Transformer to achieve performance comparable to the state-of-the-art. Second, we formulate a simple yet effective pipeline dubbed \textit{Pix4Point} that allows harnessing Transformers pretrained in the image domain to enhance downstream point cloud understanding. This is achieved through a modality-agnostic Transformer backbone with the help of a tokenizer and decoder specialized in the different domains. Pretrained on a large number of widely available images, significant gains of PViT are observed in the tasks of 3D point cloud classification, part segmentation, and semantic segmentation on ScanObjectNN, ShapeNetPart, and S3DIS, respectively.  Our code and models are available at \url{https://github.com/guochengqian/Pix4Point}.
\end{abstract}

\section{Introduction}
\label{sec:intro}

Point clouds are one of the most essential 3D representations, with broad applications in robotics, autonomous driving, medical imaging analysis, \etc. 
However, point clouds are quite expensive to acquire, with data cleaning and annotation costs being even higher \cite{urban3d}. 
Thus, despite the impressive performance gains made possible by deep learning \cite{krizhevsky2012alexnet,he2016resnet,qi2017pointnet,qi2017pointnet++,Egoloc}, \textit{their need for massive labeled training data limits possible applications of deep neural networks in point clouds}.
For instance, while Transformers \cite{vaswani2017attention} 
have achieved great success in natural language processing and computer vision \cite{devlin2018bert, dosovitskiy2021vit}, they result in relatively poor performance in point cloud understanding \cite{yu2022pointbert, pang2022pointmae}.
As a concrete example, Standard Transformer\footnote{Standard Transformer refers to a neural architecture using the original Transformer \cite{vaswani2017attention} as the backbone, \eg BERT \cite{devlin2018bert}, ViT \cite{dosovitskiy2021vit}.} only reaches $77.2\%$ overall accuracy in real-world point cloud classification \cite{uy2019scanobjectnn}, meanwhile, state-of-the-art convolutional neural network PointNeXt \cite{qian2022pointnext} attains much better accuracy ($87.7\%$).

\begin{figure}[t]
\centering
\includegraphics[page=1, trim=0 2.1inch 3inch 0, clip, width=1.0\columnwidth]{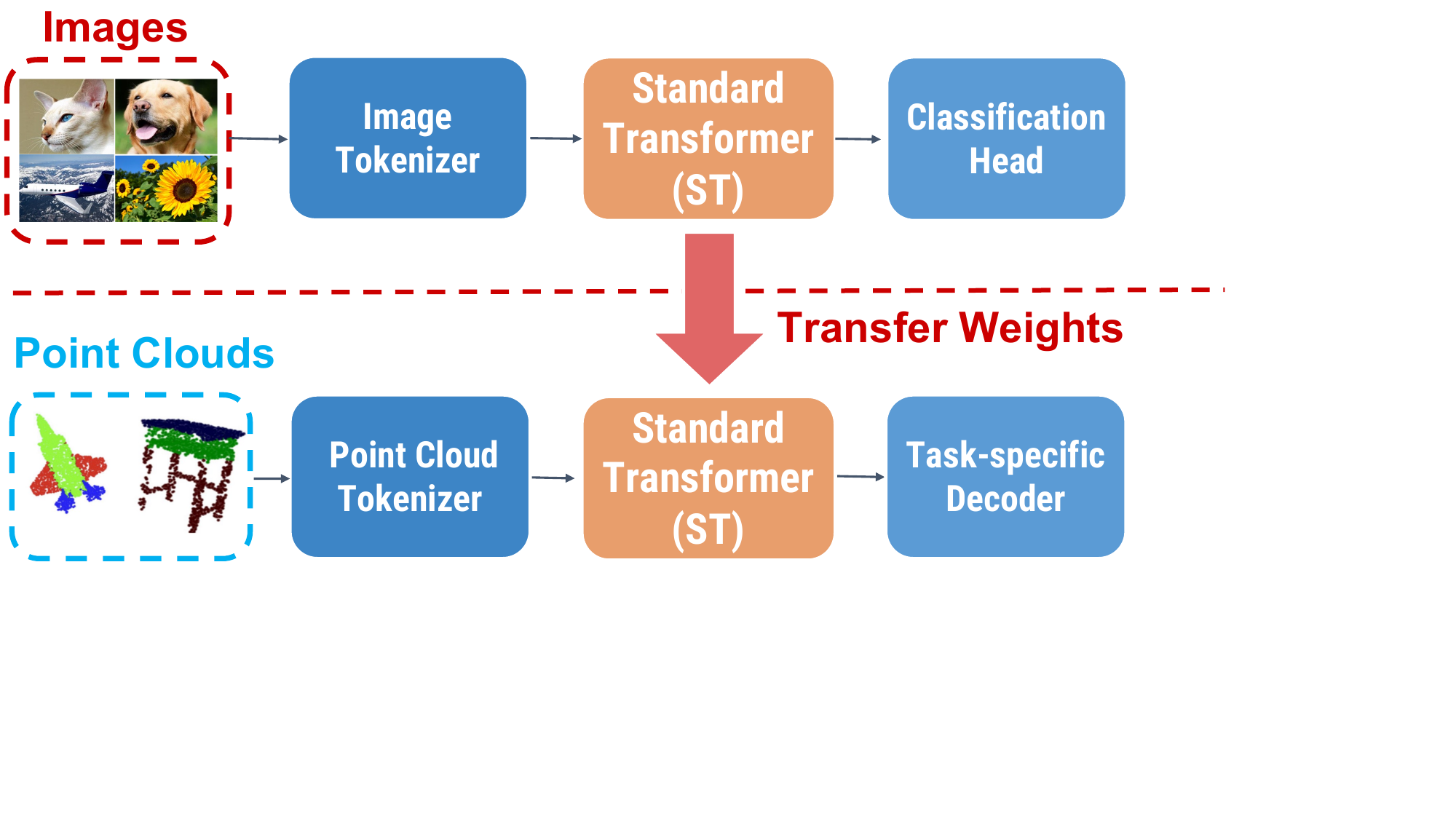}
\caption{\textbf{Image-Pretrained Transformer for Point Clouds.} Standard Transformers pretrained on images can be applied directly to point clouds and improve performance on a variety of 3D tasks including classification, segmentation, and part segmentation.
}
\label{fig:teaser}
\end{figure}

Despite its low performance, the study of Transformer is of practical importance.
Transformer can process data from all modalities, including language, images, and point clouds \cite{vaswani2017attention,dosovitskiy2021vit,yu2022pointbert,chen2022visualgpt}, working as an universal computation engine \cite{UniversalEngines}. Recent work \cite{reed2022gato} also shows that Transformer can serve as a generalist agent for more than $600$ tasks across domains. This cross-modality ability that allows for exchanging and leveraging knowledge across domains is unique to Transformers rather than CNNs or hybrid models. 
Unfortunately, previous studies on point cloud Transformers diverge from this Standard Transformer trend. They propose to integrate convolutions into Transformers such as PCT \cite{guo2021pct} or conducting self-attention locally like Point Transformer \cite{zhao2021pointtransformer}. 
These approaches make the Transformers specialized for point clouds rather than being standard. 
\textit{Empowering Standard Transformers to perform well in point cloud understanding is important but remains under-studied.}
In this work, we notice the main issue for training 3D Transformer: Transformer demands massive training data but the number of 3D points clouds is limited. 
This data issue can be solved from two perspectives:

\noindent\textbf{(1)} \textbf{Inductive bias.} Transformer's dependence on data stems from the fact that Transformer does not include inductive biases, \eg translation invariance and locality as in CNNs. These inductive biases can be added through architectural innovation such as integrating more convolutions. 
As Transformer backbone should remain unchanged, we focus our study on the tokenizer. 
Inspired by related work in image recognition \cite{xiao2021early} and point cloud completion \cite{yu2021pointr}, we utilize a progressive tokenizer that gradually projects the input point clouds into tokens. However, this progressive tokenizer mainly introduces locality without translation and rotation invariance. 
Therefore, we propose a relative progressive tokenizer that uses relative positions and relative features as input. 
Our relative progressive tokenizer significantly improves Transformer's performance in point cloud tasks.

\noindent\textbf{(2)} \textbf{Cross-modality pretraining. }
3D point clouds are expensive to acquire, but other rich domains (\eg images) could be exploited for 3D point cloud tasks.
In contrast to point clouds, 2D images are much cheaper and easier to collect/annotate. For comparison, ModelNet \cite{wu20153d}, the widely used point cloud classification dataset, consists of only $12,311$ CAD models. Meanwhile, ImageNet \cite{deng2009imagenet} is over $100$ times larger with more than one million ($1,331,167$) images. 
Therefore, we propose \textit{Pix4Point}, which is a simple yet effective pipeline that allows utilizing an image-pretrained Transformer to understand point clouds in their native format. As illustrated in \figlabel \ref{fig:teaser}, Pix4Point pretrains Transformer backbone on a large amount of tokenized (\ie patchified) images (\eg from ImageNet \cite{deng2009imagenet}) and then finetunes this image-pretrained Transformer in point-cloud tasks with tokenized point clouds as input. 
Pix4Point directly leverage the weights learned in the image domain to improve the performance for 3D point cloud tasks without any change in the network's backbone and process point clouds in their native formats without projections. 

\noindent\textbf{Contributions:}
\textbf{(i)} We present a new Transformer-based network called Point Vision Transformer (PViT). PViT improves Transformer on point clouds \cite{yu2022pointbert} with a \textbf{r}elative \textbf{p}rogressive tokeizer that gradually projects a given point cloud into tokens by graph convolutions. 
\textbf{(ii)} We propose Pix4Point, a simple yet effective framework that facilitates image pretraining for point cloud understanding. Our method enables significant performance improvements of Transformer in 3D tasks on various benchmarks: 
ScanObjectNN \cite{uy2019scanobjectnn}, ShapeNetPart \cite{shapenet2015}, and S3DIS \cite{armeni2016s3dis}.
\section{Related Work}

\minorsection{Point Cloud Networks} 
Due to breakthrough progress in deep learning technology \cite{krizhevsky2012alexnet,he2016resnet,ispnet,qi2017pointnet}, current prevailing point cloud processing methods are entirely data-driven and consist of three main approaches: view-based \cite{su15mvcnn,hamdi2021mvtn,goyal2021simpleview}, voxel-based \cite{maturanaVoxNet3DConvolutional2015,graham2018sparseconv,choy2019minkowski}, and point-based~\cite{qi2017pointnet,qi2017pointnet++}.
Among these three approaches, point-based methods, which directly take point clouds as input and process in their unstructured format, are receiving increasing attention, since there is no information loss. 
PointNet~\cite{qi2017pointnet}, the pioneering work of point-based methods, proposes to model the permutation invariance of points by restricting feature extraction to be pointwise. PointNet++~\cite{qi2017pointnet++} improves PointNet by further capturing local geometric structures. Most current point-based methods focus on the design of local modules \cite{wang2019dgcnn,li2018pointcnn,wu2019pointconv,thomas2019kpconv,li2019deepgcns,li2021deepgcn,deepergcn,qian2021pugcn,qian2021assanet,qian2022pointnext,zhao2021pointtransformer}. In this work, we also pay attention to point-based methods, specifically, Transformer-based networks \cite{vaswani2017attention}, which have been proposed as a new paradigm for processing point clouds \cite{guo2021pct,zhao2021pointtransformer,yu2022pointbert}.

\minorsection{2D to 3D Transfer Learning}
A straightforward means of transferring 2D knowledge to 3D is to use view-based methods. As a example, PointCLIP \cite{zhang2022pointclip} showed that a pretrained CLIP \cite{radfordLearningTransferableVisual2021} model can be used directly for zero-shot point cloud classification via image projection. More advanced methods leverage point-pixel correspondences \cite{liu20213d--2d,hamdi2021voint,wang2022p2p,Sparf} between point clouds and multi-view projected images. 
More recently, Image2Point \cite{xu2021image2point} presents a kernel inflation technique that expands kernels of a 2D CNN into 3D kernels and applies them to voxel-based point cloud understanding. 
However, multi-view methods can lead to non-ideal views and a loss of geometric information \cite{strike,hamdi2021mvtn}, while kernel inflation only shows insignificant performance gains given the gap between 2D and 3D kernels \cite{xu2021image2point}. 
In this work, we present a novel pipeline that can directly utilize the Transformer pretrained on images for 3D point cloud tasks, without image projection or kernel inflation.

\minorsection{Self-Supervised Learning on Point Clouds}
One prominent way to tackle the need for large labeled data is to rely on self-supervised learning (SSL) \cite{chen2020simple,caron2021dino}. Previous works in SSL for point clouds rely on pretext tasks \cite{Achlioptas2018generativepointcloud,sauder2019reconstruct} or auto-encoders \cite{Li2018SONetSN,Achlioptas2018generativepointcloud,wang2021occo,yan2022implicit}.
PointContrast \cite{Xie2020pointcontrast} proposes to generate two views of the point cloud with random transformations and ask the neural network to minimize the distance between matched points.
On the other hand, Point-BERT \cite{yu2022pointbert} and Point-MAE \cite{pang2022pointmae} propose masked auto-encoders to learn useful representations for 3D classification. 
Our method does not count on the availability of large amounts of point cloud data, as is usually needed in SSL. Instead, we focus on how to utilize already available image data for pertaining point cloud networks.

\minorsection{Transformer}
Based on its multi-head attention block, the Transformer \cite{vaswani2017attention} is believed to be the most successful architecture for natural language process (NLP). 
Vision Transformer (ViT) explores the direct application of this Transformer on image patches to solve vision tasks \cite{dosovitskiy2021vit}.
There is also an increasing interest in Transformer-like architectures for point cloud processing. 
Unfortunately, Transformers \cite{vaswani2017attention, dosovitskiy2021vit} cannot achieve results comparable to their convolutional counterparts in point cloud tasks \cite{yu2022pointbert}, mostly due to the small scale of point cloud data. This leads to a boom of specially designed variants of Transformer-based models \cite{yanPointasnlRobustPoint2020,hertzPointgmmNeuralGmm2020,guo2021pct,zhao2021pointtransformer,lai2022stratified}.
In contrast to these previous works that specialized the Transformer architecture, we maintain the same Transformer architecture and study means of improving its performance, as we believe Transformer can be used as an universal computation engine \cite{UniversalEngines,reed2022gato,Girdhar_2022_omnivore} and thus keeping its original architecture is more promising than the design of specialized Transformers.


\begin{figure*}[t]
\centering
\includegraphics[page=2, trim=0 2.8cm 0 0, clip, width=1.0\linewidth]{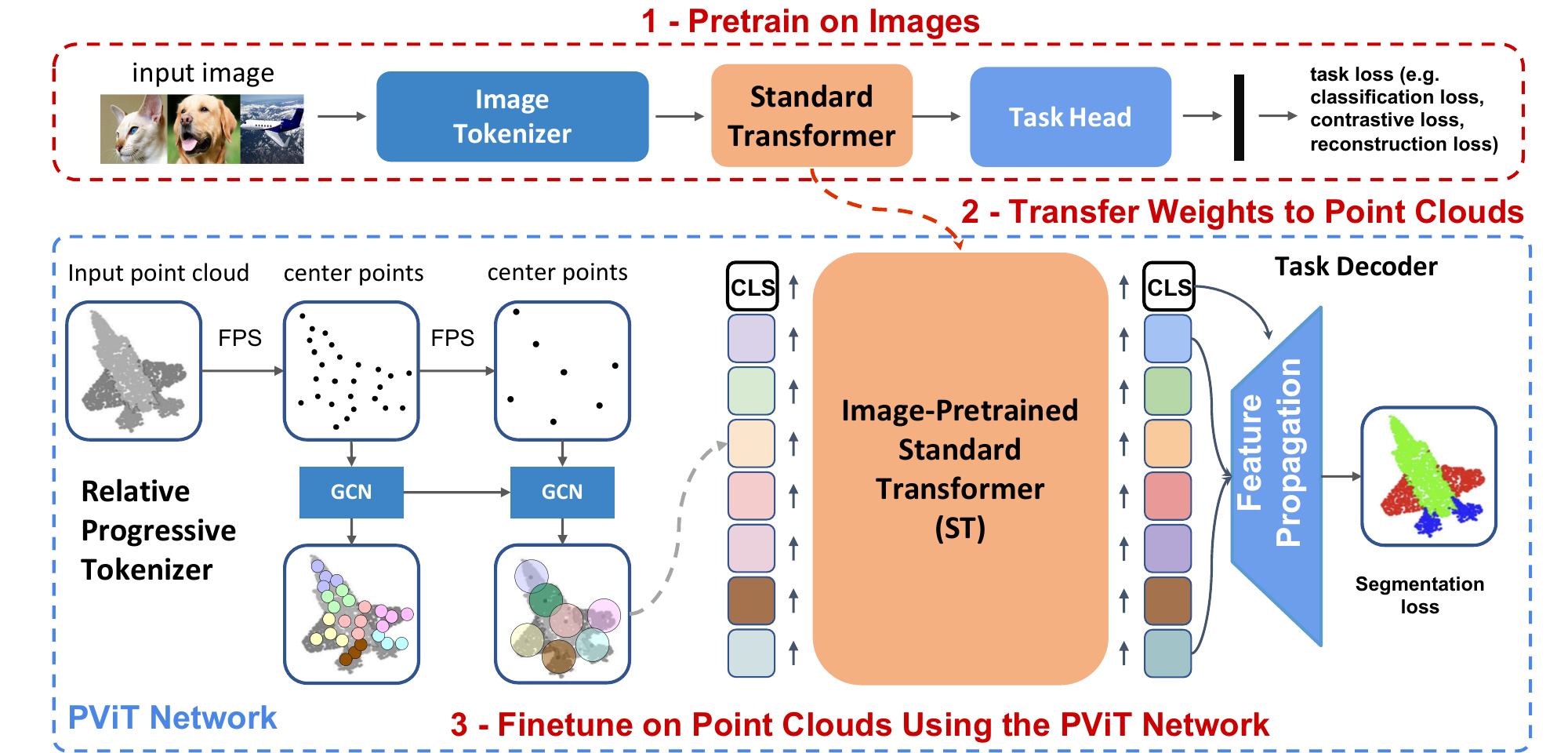}
\caption{ \textbf{Pix4Point Pipeline and PViT Network}. 
Pix4Point is composed of three stages: (1) image pretraining, (2) weight transferring, and (3) PViT downstream finetuning. PViT first projects the input point cloud into point tokens through the Relative Progressive Tokenizer $\mathbf{t}$, passes the tokens into the image-pretrained Transformer backbone $\mathbf{F}$, and then generates task outputs through the task-specific decoder $\mathbf{g}$. The parameters of $\mathbf{t}, \mathbf{F}$, and $\mathbf{g}$ are optimized jointly in the finetuning stage. Refer to sec. \ref{sec:network} and sec. \ref{sec:pix4point_pipeline} for the detailed architecture and pipeline, respectively.
}
\label{fig:pipeline}
\end{figure*}
\section{Methodology}
This section first introduces a new Transformer-based network called \textit{PViT} that improves the performance of Transformer for point cloud understanding. 
Then, we present \textit{Pix4Point}, a practical framework that utilizes plentiful available images to help point cloud tasks. PViT and Pix4Point are illustrated in \figlabel{\ref{fig:pipeline}}.

\subsection{PViT Network}
\label{sec:network}
Inspired by ViT \cite{dosovitskiy2021vit}, Yu \etal \cite{yu2022pointbert} is the pioneer to use Transformer \cite{vaswani2017attention} for point cloud understanding. They presented a point tokenizer that groups the input point cloud into local patches by a single farthest point sampling (FPS) operation to sample center points, a k-nearest neighbor (kNN) layer to group neighbors, and a mini-PointNet \cite{qi2017pointnet} to project local patches to point tokens. 
This extension of ViT \cite{dosovitskiy2021vit} to point clouds achieves poor performance in several 3D benchmarks. 
A follow-up work named Point-MAE \cite{pang2022pointmae} used the same vanilla Transformer, but relied on a better SSL method to improve the performance. 
In this work, we propose a new Transformer model called Point Vision Transformer (PViT) that can achieve comparable performance to the state-of-the-art. PViT consists of three modules: a \textit{tokenizer}, an \textit{Transformer backbone}, and a \textit{task decoder}. 
While the backbone architecture is the same as Transformer, the tokenizer and the decoder are specialized to the domain and task.

\minorsection{Tokenizer $\mathbf{t}$: Relative Progressive Tokenizer}
To improve the performance of point cloud Transformer model, the tokenzier can be designed in a way to include as many inductive biases as possible. 
Consider input point cloud $\mathcal{X} \in \mathbb{R}^{N \times (3+C_{in})}$ representing $N$ points with $C_{in}$ input features (\eg colors, normals), the tokenizer $\mathbf{t}$ projects the input point cloud into a sub-cloud (a smaller number of points but in a higher dimensional space). Our \textit{Relative Progressive Tokenizer} progressively tokenizes point clouds through multiple stages with $\mathbf{t}$: $\mathbb{R}^{N \times (3+C_{in})} \rightarrow \mathbb{R}^{\frac{N}{4} \times \frac{C}{2}} \rightarrow \mathbb{R}^{\frac{N}{16} \times C}$. At each stage,  the tokenizer $\mathbf{t}$ uses FPS to sample a subset of points with a fixed sample ratio ($1/4$), queries neighbors for each downsampled point, and leverages graph convolutions to abstract features to sub-clouds. Being progressive, our tokenizer consists of two FPS stages rather than one in Point-BERT \cite{yu2022pointbert} and Point-MAE \cite{pang2022pointmae}. The graph convolution in each stage of $\textbf{t}$ is defined as Eq. \ref{eqn:tokenizer}:
\begin{equation}\label{eqn:tokenizer}
\begin{split}
\mathbf{x}_{ij}
=h_{1\mathbf\Theta}\left([\mathbf{p}_j - \mathbf{p}_i;\mathbf{x}_j - \mathbf{x}_i]\right),\\
\mathbf{x}_{i} =h_{2\mathbf\Theta}\left([\mathbf{x}_{ij};\underset{j:(i, j) \in \mathcal{N}}{\operatorname{MAX}}{\mathbf{x}_{ij}}]\right);
\end{split}
\end{equation}
where $\mathbf{p}_i$ and $\mathbf{x}_i$ denote the coordinates and features of the $i$-th center point. The index $j$ defines the $j$-th neighbor of point $i$. The proposed graph convolution extracts features for each center point from the \textit{relative positions} and \textit{relative features} of its neighbors. The maxpooled features inside each neighborhood, ${\operatorname{MAX}}_{j:(i, j) \in \mathcal{N}}{\mathbf{x}_{ij}}$, enrich the tokens with global information from the local groups. 
Both $h_{1\mathbf\Theta}$ and $h_{2\mathbf\Theta}$ are two-layer MLPs equipped with batch normalization and ReLU.

\minorsection{Decoder $\mathbf{g}$: Feature Propagation with Global Representation Appending}
The goal of decoder $\mathbf{g}$ is to yield the desired output for each task.  For segmentation, we introduce Feature Propagation, a successful and widely used decoder design from PointNet++ \cite{qi2017pointnet++}, into Transformer. Feature Propagation gradually interpolates the feature map and concatenates it with point features from the tokenizer to produce individual representations for each point. Global representations, \ie the [CLS] token and the global maxpooled non-class tokens, are appended to each point. The global representation appending for every point $\mathbf{x}_{i} $ in the segmentation task can be formalized as:
$
[\mathbf{x}_i;\operatorname{MAX}\left(\mathbf{x}\right);\operatorname{[CLS]}]
$.
In the case of classification, the global representation appending is formalized as: 
$
[\operatorname{MAX}\left(\mathbf{x}\right);\operatorname{[CLS]}]
$.
This global representation appending technique enriches global information for both classification and segmentation and improves performance by non-trivial margins.

\minorsection{Comments}
The progressive tokenizer is inspired by the related work in image classification ViT$_C$ \cite{xiao2021early}, which finds that using a progressive tokenizer introduces locality into the Transformer model and thus offers performance gains.
The progressive tokenizer is also used in the representative point cloud completion work PointTr \cite{yu2021pointr}. 
In this work, we bring this existing technique into point cloud recognition. 
However, we find that using a progressive tokenizer alone only leads to a marginal performance improvement. 
We further apply relative position and relative feature learning into the progressive tokenizer to introduce translation and rotation invariance. 
We note that learning from relative position and relative feature also exists in the point cloud domain \cite{qin2022geometric,ran2022surface}.
Our novelty lies in the new relative progressive tokenizer that seamlessly combines progressive tokenizer with relative position and feature learning. We are also the first to study these two techniques in Transformer-based point cloud recognition.

\subsection{Pix4Point Pipeline}
\label{sec:pix4point_pipeline}
To further tackle the sparsity of available 3D data, we argue that images are easily available and could be exploited for point cloud understanding. To this end, we propose Pix4Point, a framework that enables improving PViT through image pretraining.
We illustrate the pipeline of Pix4Point in Fig. \ref{fig:pipeline}. It consists of three stages: (1) pretraining on images, (2) transferring image-pretrained weights to PViT, and (3) finetuning PViT on point cloud tasks. 

\minorsection{Pretraining Stage}
The goal of this stage is to utilize a large number of public images to pretrain the Transformer backbone of PViT. By default, we pretrain \textit{ViT-S} using the self-supervised method MAE \cite{he2022mae} on \emph{ImageNet-1K} \cite{deng2009imagenet} dataset. Note that our Pix4Point is generally agnostic of the  pretraining strategy. Besides the aforementioned self-supervised pretraining, there are many other applicable strategies, \eg supervised training through DeiT \cite{touvron2021deit}. See Sec. \ref{sec:ablation} for the effects of different pretraining strategies.

\minorsection{Weight Transfer Stage}
The goal of weight transfer stage is to leverage the image-pretrained weights in the downstream point cloud network. 
This stage in Pix4Point is as simple as using the weights of the image-pretrained Transformer as an initialization for PViT backbone. This is possible because the exact same Transformer backbone (\eg ViT-S backbone) is used in both the image pretraining and the downstream point cloud tasks.

\minorsection{Finetuning Stage}
The finetuning stage finetunes the image-pretrained Transformer backbone $\mathbf{F}$ for point cloud tasks. The tokenizer $\mathbf{t}$ and the task-specific decoder $\mathbf{g}$ of PViT are jointly trained from scratch during this fine-tuning stage. The loss for Pix4Point can be described as follows:
\begin{equation}
\begin{aligned} 
\argmin_{\boldsymbol{\theta}_{\mathbf{t}},\boldsymbol{\theta}_{\mathbf{F}}, \boldsymbol{\theta}_{\mathbf{g}}} &~~\sum_{i}^{B} L~\Big( \mathbf{g} \big(  \mathbf{F}\left( \mathbf{t}\left({\mathcal{X}}\right)\right)  \big)_i ~,~\mathbf{y}_i \Big), 
\label{eq:pipeline}
\end{aligned} 
\end{equation}
where $\{\mathbf{y}_i\}_{i=1}^B$ defines the labels of a mini-batch of input point clouds $\mathcal{X}$ and $L$ is a Cross-Entropy (CE) loss. 

\noindent\textbf{Why does Pix4Point Work?}
The 2D tokenizer and the 3D tokenizer project the input images and the point clouds into a shared token space $\mathcal{R}^C$, respectively, where each token from 2D or 3D is embedded as a vector of the same length $C$. 
Despite the domain shift between 2D and 3D and owing to the diverse large-scale image data in the pretraining stage, the Transformer backbone has learned to transform and mix the information of tokens in this shared token space. 
Therefore, it only takes a few finetuning epochs to tune Transformer and optimize the domain-specific tokenizer for the downstream point cloud tasks. Interestingly, we show  that even when the image pretrained Transformer backbone is kept frozen, our proposed pipeline can still adequately recognize point clouds by training only the tokenizer and decoder in the downstream task (see Sec. \ref{sec:ablation}).
This verifies our hypothesis that how to transform and mix information in the shared token space has been learned during the image pretraining stage.

\minorsection{Compare with Image2Point}
Image2Point \cite{xu2021image2point} focuses on CNNs and expands the image pretrained 2D kernels into 3D kernels through kernel inflation. In contrast, our Pix4Point focuses on Transformer and simply uses the image-pretrained Transformer as initialization for downstream point cloud tasks. 
Despite this major difference, one can still think Pix4Point is a special case of Image2Point by using the identity function as kernel inflation. 
Here, we highlight that studying
such an extreme case carefully of practical importance due to the increasing interest of using Transformers.
\section{Experiments and Results
}\label{sec:exp}
\begin{figure*}[t]
\centering
\includegraphics[page=3, trim=0cm 0cm 0cm 0cm, clip, width=1.0\textwidth]{src/pipeline.pdf}
\includegraphics[trim=0cm 1.5cm 0cm 0.2cm, clip, width=1.0\textwidth]{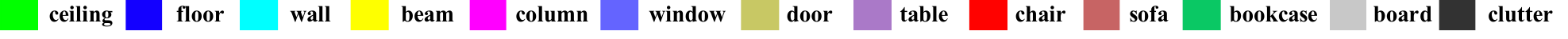}
\vspace{-1.5em}
\caption{\textbf{Qualitative Results of PViT on S3DIS Area 5.} PViT with image pretraining ($4^{th}$ column) achieves more precise segmentation results than PViT trained from scratch ($3^{rd}$ column) and Point-BERT \cite{yu2022pointbert} ($2^{nd} column$).
}
\label{fig:qualitative}
\end{figure*}

We conduct extensive experiments on various 3D benchmarks, including \emph{S3DIS} \cite{armeni2016s3dis} segmentation, \emph{ShapeNetPart} \cite{shapenet2015} segmentation, and \emph{ScanObjectNN} \cite{uy2019scanobjectnn} classification.
We highlight the comparison among Transformer-based methods: PViT, PViT+Pix4Point, Transformer \cite{yu2022pointbert}, Point-BERT \cite{yu2022pointbert}, and Point-MAE \cite{pang2022pointmae}. All of these methods use the same Transformer backbone (ViT-S) with $12$ self-attention blocks, $384$ channel size, and $6$ heads for all tasks.  Transformer \cite{yu2022pointbert} uses the same network in Point-BERT trained from scratch without self-supervised learning. 

\subsection{Training Setup}
\minorsection{Pretraining setup} 
We pretrain ViT-S \cite{dosovitskiy2021vit} using MAE \cite{he2022mae} on \emph{ImageNet-1K}~\cite{deng2009imagenet} in a self-supervised fashion. The final weights of ViT-S backbone (epoch 400) are used to initialize the backbone of our PViT network.

\minorsection{Finetuning Setup}
For segmentation on S3DIS, PViT is trained using the Cross-Entropy loss with $0.2$ label smoothing, AdamW optimizer \cite{loshchilov2019adamw} with a learning rate of 1e-4, a cosine learning rate scheduler, warmup epochs of 10, a weight decay of 1e-4, a batch size of $8$, and for $600$ epochs. 
Data augmentation includes rotation, scaling, jittering, color auto-contrast, and color dropping.
For part segmentation on ShapeNetPart, PViT is trained similarly as on S3DIS using $2,048$ points as input. The difference is: the models are trained for $500$ epochs with a learning rate of 5e-4 and we drop normals instead of colors and do not perform color auto-contrast, since there is no color information in ShapeNetPart.
For classification on ScanObjectNN, PViT is trained similarly as on ShapeNetPart. A weight decay of $0.05$ and $300$ epochs with a batch size of $32$ are used. The number of input points is set to 1,024, where the points are randomly sampled during training and uniformly sampled during testing. 
In all experiments, the best model in the validation set is selected for testing.
Note the same training setup is used for both PViT trained from scratch and PViT+Pix4Point.

\begin{table}[t]
\centering
\caption{\textbf{Semantic Segmentation on S3DIS Area 5.}
Methods are divided into two categories. \emph{Bottom}: Transformers. \emph{Top}: other methods. All methods in the bottom category use the same ViT-S backbone. 
PViT is trained from scratch, while PViT+Pix4Point is trained with an \textit{image-pretrained} backbone. 
PViT outperforms Standard Transformer \cite{yu2022pointbert} due to the improved tokenizer and decoder. 
With image pretaining, Pix4Point enables PViT to achieve comparable performance to the \sota.
\hl{Green} color highlights improvements over Transformer. 
}
\label{tab:s3dis_area5}
\vspace{-0.5em}
\resizebox{0.95\linewidth}{!}{
\begin{tabular}{ 
l | l l 
}
\toprule[1pt]
\textbf{Method}
&   \textbf{mIoU} 
&   \textbf{mAcc} 
\\
 & (\%) & (\%)  
 \\
\midrule
PointNet~\cite{qi2017pointnet} 
& 41.1 & 49.0 
\\ 
DeepGCN~\cite{li2019deepgcns}
& 52.5& -  
\\ 
PointNet++~\cite{qi2017pointnet++} 
& 53.5& - 
\\ 

PVCNN~\cite{liu2019point} 
 & 59.0& -
\\ 

PCT~\cite{guo2021pct} 
& 61.3& 67.7
\\ 



ASSANet-L~\cite{qian2021assanet} 
& 66.8& - 
\\ 

KPConv~\cite{thomas2019kpconv} 
& 67.1& 72.8 
\\ 

Point Transformer~\cite{zhao2021pointtransformer} 
& {70.4}& {76.5} 
\\ 
PointNeXt~\cite{qian2022pointnext} 
& 70.5& 76.8 
\\
\midrule
Transformer~\cite{yu2022pointbert}  & 60.0 & 68.6 
\\
Point-BERT~\cite{yu2022pointbert}  & 60.8 & 69.9 
\\
\gr \textbf{PViT}
&	\reshl{64.4}{+}{4.4} 
&	\reshl{69.9}{+}{1.3} 
\\
\gr \textbf{PViT+Pix4Point (Ours)}
&   \reshl{69.6}{+}{9.6}
&   \reshl{75.2}{+}{6.6}
\\
\bottomrule[1pt]
\end{tabular}}
\end{table}
\subsection{3D Semantic Segmentation}
\minorsection{Dataset}
S3DIS~\cite{armeni2016s3dis} (Stanford Large-Scale 3D Indoor Spaces) dataset provides instance-level semantic segmentation for $6$ large indoor areas that cover $271$ rooms and $13$ semantic categories. Each room contains averagely $795$K points, making S3DIS a challenging segmentation benchmark.
Following common practice, we leave area 5 for testing and use the rest for training.

\minorsection{Results}
We show the quantitative results of PViT compared to \sota methods on S3DIS area 5 in Tab. \ref{tab:s3dis_area5}. \textbf{(1)} We find that Standard Transformer \cite{yu2022pointbert} achieves much lower performance than CNN-based alternatives such as KPConv \cite{thomas2019kpconv}, and specialized Transformer networks for point cloud understanding \eg PCT \cite{guo2021pct} and Point Transformer \cite{zhao2021pointtransformer}. 
This occurs because Transformer is known to be harder to optimize.
\textbf{(2)} Our improved Transformer-based model (PViT) outperforms Transformer by a significant margin ($+4.4$ mIoU). This shows the effects of our proposed tokenizer and Feature Propagation decoder with global representation appending. 
\textbf{(3)} Most importantly, results demonstrate that image pretraining using Pix4Point impressively further improves PViT by $5.2\%$ in mean IoU (mIoU) and $5.3\%$ in mean accuracy (mAcc). This achievement is simply due to initializing the Transformer backbone with an image-pretrained Transformer. With image pretraining, we show that an Transformer-based network can perform much better than representative 3D methods including PointNet++ \cite{qi2017pointnet++}, PVCNN \cite{liu2019pvcnn}, and KPConv \cite{thomas2019kpconv}. 
Compared to Point Transformer \cite{zhao2021pointtransformer}, a highly optimized local Transformer-based network with a hierarchical structure, 
PViT with image pretraining can achieve comparable performance to it while using only the Standard Transformer as the backbone. 
These observations verify our argument: progressive tokenizer and global appending improve the performance of Transformer, and image pretraining can be beneficial for point cloud understanding. 

\begin{table}[tb]
\small
\centering
\caption{\textbf{Part Segmentation on ShapeNetPart.}}
\label{tab:shapenetpart}
\vspace{-0.5em}
\resizebox{0.95\linewidth}{!}{%
\begin{tabular}{l|ll}
\toprule
\textbf{Method}  & \textbf{Ins.\ mIoU}  & \textbf{cls. \ mIoU}   
\\ 
\midrule
PointNet~\cite{qi2017pointnet} & 83.7 & 80.4 
\\
 PointNet++~\cite{qi2017pointnet++} & 85.1 & 81.9
 \\

DGCNN~\cite{wang2019dgcnn}  & 85.2 & 82.3
\\
ASSANet-L~\cite{qian2021assanet} & 86.1 & -
\\
PointMLP~\cite{ma2022pointmlp} & 86.1 & 84.6 
\\
KPConv~\cite{thomas2019kpconv}& 86.4 & 85.1 
\\

PCT~\cite{guo2021pct} & 	86.4 &- 
\\ 
Point Transformer~\cite{zhao2021pointtransformer} & 86.6 & 83.7 
\\ 

StratifiedFormer~\cite{lai2022stratified} & 86.6 &  85.1
\\ 
PointNeXt~\cite{qian2022pointnext} & 87.0 & 85.2 
\\
\midrule
Transformer~\cite{yu2022pointbert} & 85.1 & 83.4 
\\
Point-BERT~\cite{yu2022pointbert} & 85.6 & 84.1 
\\
Point-MAE~\cite{pang2022pointmae} & 86.1 & 84.2 
\\
\gr \textbf{PViT} &  \reshl{85.7}{+}{0.6} & \reshl{83.7}{+}{0.3} 
\\
\gr \textbf{PViT+Pix4Point (Ours)} & \reshl{\textbf{86.8}}{+}{1.7} & \reshl{\textbf{85.6}}{+}{2.2} 
\\
\bottomrule
\end{tabular}
}
\end{table}

\minorsection{Qualitative Results}
Fig. \ref{fig:qualitative} shows some qualitative results of PViT and PViT+Pix4Point. The latter yields more precise segmentation maps than the former. Red circles show that image pretraining helps PViT successfully segment the boards ($1^{st}$ and $3^{rd}$ rows), the door ($2^{nd}$ row), and the columns ($2^{nd}$ and $3^{rd}$ rows). PViT also achieves better qualitative results than Point-BERT \cite{yu2022pointbert} self-supervised on point clouds.

\subsection{3D Part Segmentation}
\minorsection{Dataset}
ShapeNetPart~\cite{shapenetparts} is a  richly annotated 3D dataset of 16 shape categories selected from the ShapeNet dataset,  annotated with part-level semantic labels. It consists of $16,880$ models, $2-6$ parts for each category, and $50$ part labels in total. Pix4Point learns a single model using a single shared head module to predict all parts. 

\minorsection{Results} The performance of PViT on ShapeNetPart is reported in Tab. \ref{tab:shapenetpart}. 
\textbf{(1)}
Owing to the improved point cloud tokenizer and the global representation appending in the decoder, PViT trained from scratch outperforms Transformer by $0.6$ instance mIoU and $0.3$ class mIoU.
\textbf{(2)}
Image pretraining further gains $+1.1$ in instance mIoU and $+1.9$ in terms of class mIoU. With Pix4Point image pretraining, PViT achieves $86.8$ instance mIoU, outperforming the Transformer-based networks such as Point Transformer \cite{zhao2021pointtransformer}, and being close to the state-of-the-art PointNeXt \cite{qian2022pointnext}.

\begin{table}[tb]
\small
\centering
\caption{\textbf{3D Classification on ScanObjectNN PB\_T50\_RS.}
}
\label{tab:cls_scanobjectNN}
\vspace{-0.5em}
\resizebox{0.8\linewidth}{!}{
\begin{tabular}{l|lll}
\toprule
\textbf{Method}& \textbf{OA}  & \textbf{mAcc}   
\\
 & (\%) & (\%)  
 \\ 
\midrule
PointNet~\cite{qi2017pointnet}
& 68.2& 63.4 
\\
PointNet++~\cite{qi2017pointnet++} & 77.9& 75.4
\\
PointCNN~\cite{li2018pointcnn}     & 78.5& 75.1 
\\
DGCNN~\cite{wang2019dgcnn}         & 78.1& 73.6 
\\  
PointMLP~\cite{ma2022pointmlp}     & 86.4& 83.9 
\\ 
PointNeXt~\cite{qian2022pointnext} & 87.7 & 85.8 
\\ 
\midrule
Transformer~\cite{yu2022pointbert} &  
77.2 & - 
\\
Point-BERT~\cite{yu2022pointbert}  & 83.1& - 
\\
Point-MAE~\cite{pang2022pointmae}   & 85.2&- 
\\
\gr PViT   & \reshl{85.7}{+}{8.5}  & 83.5 
\\ %
\gr \textbf{PViT+Pix4Point}  & \reshl{\textbf{87.9}}{+}{10.7} &\textbf{86.7} 
\\  
\bottomrule
\end{tabular}
}
\end{table}

\subsection{3D Object Classification}
\minorsection{Dataset}
ScanObjectNN~\cite{uy2019scanobjectnn} collects a total of 15,000 scanned objects for 15 classes. This real-world dataset presents challenges to classification tasks due to inherent scan noise and occlusion. We benchmark Pix4Point on the hardest variant of this dataset, PB\_T50\_RS. 

\minorsection{Results} Tab. \ref{tab:cls_scanobjectNN} shows the effectiveness of the proposed PViT in point cloud classification in the real dataset ScanObjectNN. 
\textbf{(1)} PViT trained from scratch achieves better performance than the Transformer network \cite{yu2022pointbert} as well as SSL-based methods Point-BERT \cite{yu2022pointbert} and Point-MAE \cite{pang2022pointmae}.
\textbf{(2)} Image pretraining improves the overall accuracy (OA) and the mean accuracy (mAcc) of PViT by $2.2\%$ and $3.2\%$, respectively. With Pix4Point image pretraining, PViT achieves a superior $87.9\%$ OA in ScanObjectNN classification.
\section{Ablation Study and Analysis}\label{sec:ablation}

In this section, we study the effects of pretraining strategies on the downstream point cloud tasks, including pretraining on different datasets (images or point clouds), using supervised or self-supervised pretraining. 
We further ablate the architecture design of the tokenizer and decoder for the proposed PViT network. 

\begin{table}[tb]
\centering
\caption{\textbf{Effect of Pretraining Strategies.}
We show the downstream performance of PViT on S3DIS area 5 using the ViT-S backbone pretrained by (1) SSL on ShapeNet by Point-MAE \cite{pang2022pointmae} ($2^{nd}$ row), (2) PViT supervised training from scratch through the ShapeNet part segmentation task ($3^{rd}$ row), (3) SSL on ImageNet by MAE \cite{he2022mae} ($4^{th}$ row), and (4) supervised training on ImageNet by DeiT \cite{touvron2021deit} ($5^{th}$ row). The random initialized Transformer backbone ($1^{st}$ row) is provided for reference. Network architectures (\ie PViT) and the finetuning parameters are the same for all experiments for a fair comparison. Entire Network and Frozen Backbone refer to optimizing the whole network and only the tokenizer and decoder, respectively. 
\textit{Image pertaining improves point cloud understanding in various setups, especially when the Transformer backbone is frozen.}
}
\label{tab:pretraining_methods}
\vspace{-0.5em}
\resizebox{1.0\linewidth}{!}{
\begin{tabular}{ l | l | l}
\toprule
\multirow{3}{*}{\textbf{Pretraining strategies}}
& Entire Network  & \textit{Frozen} backbone \\
& {mIoU} 
& {mIoU} 
\\
\toprule
scratch
&64.4
& 54.7 
\\
\midrule
Point-MAE 
&\reshl{66.2}{+}{1.8}
&\reshl{55.9}{+}{1.2}
\\
PViT on ShapeNet
&\reshl{65.6}{+}{1.2}
&\reshl{54.9}{+}{0.2}
\\
\midrule
\gr
\textbf{MAE}
& \reshl{\textbf{69.6}}{+}{5.2}
& \reshl{62.0}{+}{7.3} 
\\
\gr
DeiT
& \reshl{68.3}{+}{3.9}
& \reshl{\textbf{62.2}}{+}{7.5}
\\
\bottomrule
\end{tabular}}
\end{table}

\minorsection{Pretraining Strategies}
Pix4Point pretrains ViT-S using MAE on ImageNet-1K by default. Apart from this self-supervised pretraining on images, there are multiple other ways to pretrain Transformer. It can be pretrained on 3D or 2D datasets, as well as, in a supervised or self-supervised manner. Here, we compare the following pretraining strategies: (1) self-supervised pretraining on ImageNet using MAE \cite{he2022mae}; (2) supervised pretraining on ImageNet using DeiT \cite{touvron2021deit} without distillation;  (3) self-supervised pretraining on ShapeNet using Point-MAE \cite{pang2022pointmae};
(4) supervised pretraining using the PViT network in the ShapeNet part segmentation task. 
Note that all pretraining strategies share the same ViT-S backbone, finetuning the same network PViT using the same finetuning parameters for a fair comparison. 

Tab. \ref{tab:pretraining_methods} compares the finetuning performance on S3DIS of the aforementioned training strategies. 
\textbf{(1)} While finetuning the entire network, 
\textit{the proposed image-pretrained Transformer} reach mIoUs greater than $68$ for both DeiT supervised pretraining ($68.3$ mIoU) and MAE self-supervised pretraining ($69.3$ mIoU), thus,  \textit{outperforming the Transformer trained from scratch} ($64.4$ mIoU) by more than $3.9$ mIoU. 
\textbf{(2)} More interestingly, \textit{the image-pretrained Transformer also surpass the point cloud pretrained Transformer}, both supervised (PViT pretrained on ShapeNet part segmentation) and self-supervised (Point-MAE) pretrained, by non-trivial margins (over $2$ mIoU).

\begin{figure}[t]
\centering
\centering
\includegraphics[width=1.0\linewidth]{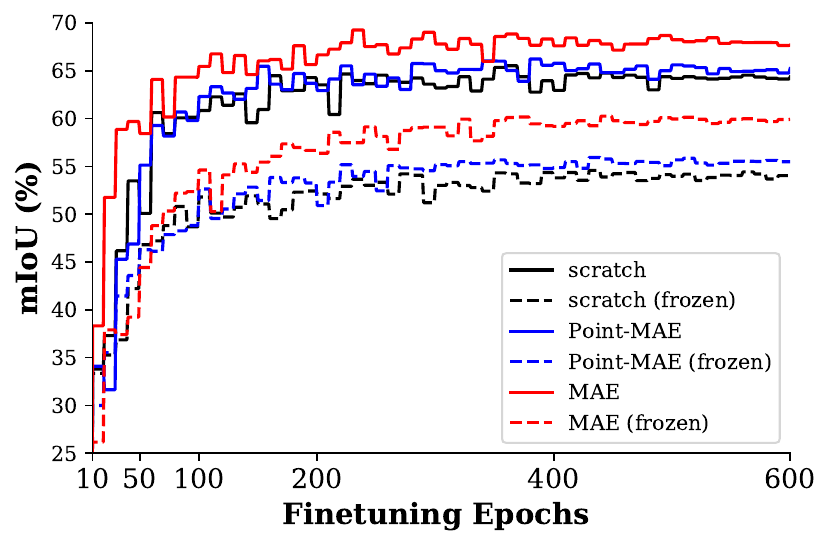}
\vspace{-2em}
\caption{ \textbf{Effect of Pretraining Strategies with and without a Frozen Backbone.} 
We show validation curves of downstream performance of PViT on S3DIS area 5 with the same backbone (ViT-S) pretrained using different strategies: from scratch, self-supervised pretraining on ShapeNet by Point-MAE, self-supervised pretraining on ImageNet-1K by MAE (Pix4Point). Results with a frozen pretrained backbone are also included for reference.
As observed, image-pretraining improves point cloud understanding more than the usual point cloud pretraining in ShapeNet.
}
\label{fig:pretrain_methods}
\end{figure}

Fig. \ref{fig:pretrain_methods} (solid lines) shows the val mIoU during fine-tuning on S3DIS of different pretraining strategies. It is observed that the \textit{image-pretrained Transformer (in red line) always outperforms the random initialized Transformer (in black line) and the point cloud pretrained Transformer (in blue line) across finetuning epochs.} These experiments clearly demonstrate the benefits of Pix4Point pretraining.

\minorsection{Effects of Image Pretraining without Finetuning the Transformer Backbone}
Previously, we showed that image pretraining improves PViT performance on various benchmarks by training the tokenizer and decoder, as well as, finetuning the pretrained Transformer. To study the impact of this backbone finetuning and know how good is the pretrained backbone, we keep the weights of the pretrained Transformer \textit{frozen} here and only update the weights of the tokenizer and decoder during finetuning. As shown in Tab. \ref{tab:pretraining_methods} and Fig. \ref{fig:pretrain_methods} (dash lines), the image-pretrained PViT still reach more than $62$\% mIoU for both DeiT supervised pretraining ($62.2$\% mIoU) and MAE self-supervised pretraining ($62.0$\% mIoU), despite the backbone is frozen, \textit{significantly outperforming the random initialized Transformer ($54.7$\% mIoU) more than $7.3$ mIoU}. Image-pretrained Transformer backbone also perform better than the same backbone pretrained on point cloud data, \ie Point-MAE ($55.9$\% mIoU) and supervised pretraining ($54.9$\% mIoU) on ShapeNet. These experiments directly demonstrate the benefits of image pretraining, since the network and finetuning parameters are the same for all methods and the pretrained Transformer is frozen during finetuning.  

\minorsection{Additive Study of the PViT Architecture}
The difference between PViT and Transformer is that our PViT uses relative progressive tokenizer and global representation appending in the decoder. Table \ref{tab:additive} conducts an ablation study to show the effects of each architectural changes on S3DIS Area 5. We start from Transformer, and gradually adds the contributed parts. The following is observed: 
\textbf{(1)} Our relative progressive tokenizer improves Transformer consistently with/without Pix4Point by more than $3\%$ mIoU. Our tokenizer also outperforms the one used in PointTr \cite{yu2021pointr} and the one without relative feature showing the importance of relative learning. 
\textbf{(2)} The global appending technique  leads to $0.8\%$ and $2.1\%$ mIoU improvements when trained from scratch and with image pretraining, respectively. 
\begin{table}[t]
\centering
\caption{\textbf{Additive Studies of the Architecture Design.}
We start our study from Transformer \cite{yu2022pointbert} (uses only relative position in tokenizer), and adds each architectural modification step by step. Test mIoUs with and without Pix4Point are compared. 
}
\label{tab:additive}
\resizebox{1.00\linewidth}{!}{\begin{tabular}{l|c|c}
\toprule
\textbf{Ablation} &\textbf{Scratch} & \textbf{Pix4Point} \\
\hline
Transformer \cite{yu2022pointbert} & 60.2 & 64.4 \\
\hline
$+$ Progressive tokenizer ([$\Delta\mathbf{p},\mathbf{x}$])&  54.6 & 56.9\\
$+$ Progressive tokenizer ([$\mathbf{x},\Delta\mathbf{x}$])&  56.6 & 58.0 \\
$+$ Progressive tokenizer ([$\mathbf{p},\Delta\mathbf{x}$], PoinTr \cite{yu2021pointr} )&  62.9	& 65.0\\
\grayrow 
$+$ Progressive tokenizer ([$\Delta\mathbf{p},\Delta\mathbf{x}$], \textbf{Ours}) & 63.6 & 67.5 \\
\hline
\grayrow 
$+$ global appending (\textbf{Our PViT}) & 64.4 & \textbf{69.6} \\
\bottomrule
\end{tabular}}%
\end{table}

\section{Conclusions and Future Work}
To have a unified Standard Transformer architecture for all modalities is the trend. However, Transformer does not work well for 3D point clouds understanding, achieving way worse performance than state-of-the-art point cloud networks.
This work goes one step further to the unified goal by introducing a progressive point cloud encoder for the Transformer backbone and an image-pretraining framework Pix4Point to boost the performance of Transformer in the task of point cloud understanding.
Pix4Point framework enables the use of publicly available 2D images to pretrain the Transformer backbone and helps the understanding of downstream 3D point clouds. 
We conclude the following: \textbf{(1)} the proposed PViT network using the progressive encoder and the global representation appending improves Transformer;
\textbf{(2)} the image-pretrained Transformer can significantly boost performance in various 3D tasks; (3) ImageNet pretraining outperforms ShapeNet pretraining. We believe that our findings will motivate future work in this direction. 

While the scope of this work is to investigate the benefits of using image-pretrained Transformers in the point cloud domain, a possible future work is to investigate \textit{a generalist vision model} that is jointly optimized across the images and 3D point clouds. Such a generalist model might be helpful for a multi-modality scenario such as autonomous driving. 

\vspace{1em}
\minorsection{Acknowledgement}
We thank Hani Itani and Jun Chen for their insightful discussions. This work was supported by the KAUST Office of Sponsored Research (OSR) through the Visual Computing Center (VCC) funding. Part of the support is also coming from the KAUST Ibn Rushd Postdoc Fellowship program.

{\small
\bibliographystyle{ieee_fullname}
\bibliography{main}
}
\appendix
\renewcommand{\thesection}{\Alph{section}}
\renewcommand{\thetable}{\Roman{table}}
\renewcommand{\thefigure}{\Roman{figure}}
\setcounter{section}{0}
\setcounter{table}{0}
\setcounter{figure}{0}

\twocolumn[{%
\centering
{\huge\bf Pix4Point: Image Pretrained Standard Transformers for 3D Point Cloud Understanding
}\\
\vspace{1em}
{\large\bf--- Supplementary Material ---}
\vspace{2em}
}]

In this appendix, we provide additional content to complement the main manuscript:
\begin{itemize}[leftmargin=1em]
\item Additional analysis on the effect of pretraining epochs;
\item Additional analysis on the effect of pretraining dataset size;
\item Detailed architecture configurations of PViT in various tasks.
\end{itemize}



\section{Additional Analysis}
We provide additional analysis on the pretraining epochs and pretraining dataset size in this supplementary material. 
We use the DeiT pretrained PViT without progressive encoder as the baseline.

\minorsection{Pretraining Epochs}
Here we study the effects of pretraining epochs on the downstream performance. We train DeiT without distillation using the entire training set of ImageNet-1K. We leverage the pretrained Transformers from different epochs and test their downstream performance in S3DIS area 5. Tab.~\ref{tab:epochs} shows that the downstream performance drops at first mainly because the Transformer does not converge to good minima yet. The pretrained Transformer starts to outperform the random initialized Transformer ($62.3$ mIoU) at around epoch $50$, gradually increases to reach peak performance ($\sim66$ mIoU) at epoch $500$, and begins to decline from epoch $600$ mostly due to overfitting image data. 

\minorsection{Pretraining Dataset Size}
Here, we ablate the size of the pretrained dataset by pretraining the ST using only a portion of ImageNet-1K. All models are trained using DeiT \cite{touvron2021deit} without distillation by 300 epochs. Tab. \ref{tab:datasize} shows that the downstream performance in point cloud segmentation on S3DIS \textit{increases with the size of the pretraining dataset}. 
The performance of Pix4Point continuously improves from $61.8$ to $65.8$. Increasing the size of the dataset from ImageNet-1K to ImageNet-21K (10 times larger) can further improve the performance by $0.8$ mIoU.

\section{Architecture}
We list the detailed configuration of the proposed PViT architecture in Tab. \ref{tab:pvit_archi}. The same Standard Transformer ($12$ layers, $6$ heads, and $384$ hidden dimensions) is used as the backbone in all tasks. The maxpooled feature among all tokens and the [CLS] are appended in the decoder to improve the segmentation and classification performance with global information. 
PViT uses a progressive tokenizer to subsample the input point cloud to $1/16$ with two downsampling stages in segmentation tasks. For classification, since the number of input points is small, PViT only downsamples once with a sampling ratio equal to 1/4.

\begin{table}[t]
\centering
\caption{\textbf{Effects of Pretraining Epochs on Downstream S3DIS Area 5 Segmentation.} Pretraining strategy: DeIT without distillation on ImageNet-1K. PViT without a progressive encoder is used as the network. 
The pretrained Transformer reduces the downstream performance before epoch $50$ due to that the backbone does not converge to a good minima yet, then outperforms the random initialized Transformer ($61.8$\% mIoU), gradually increases the performance to reach a peak ($\sim66$ mIoU) at epoch $500$, and begins to decline the pretraining improvement from epoch $600$ mostly due to overfitting image data. 
}
\label{tab:epochs}
\resizebox{1.0\linewidth}{!}{
\begin{tabular}{ l | cccccccccc}
\toprule
Epoch & 0 &  10 & 20 & 50 & 100 & 200 & 300 & 400 & 500 & 600\\
\midrule
mIoU (\%)&
$61.8$&
\cellcolor{red!10}{$59.4$}&
\cellcolor{red!10}{$60.3$}&
\cellcolor{red!10}{$61.7$}&
\cellcolor{green!10}{$63.8$}&
\cellcolor{green!10}{$64.9$}&
\cellcolor{green!10}{$65.8$}&
\cellcolor{green!10}{$66.0$}&
\cellcolor{green!10}{$\textbf{66.6}$}&
\cellcolor{red!10}{$65.8$}\\
\bottomrule
\end{tabular}}
\end{table}
\begin{table}[tb]
\centering
\caption{\textbf{Effects of Pretraining Dataset Size on Downstream S3DIS Area 5 Segmentation.} Pretraining strategy: DeIT without distillation on a portion of ImageNet-1K dataset. The downstream performance in point cloud segmentation on S3DIS increases with the size of the pretraining dataset. 
}
\label{tab:datasize}
\resizebox{1.0\linewidth}{!}{
\begin{tabular}{ l | cccccccccc}
\toprule
Portion (\%) & 0 & 10 & 20 & 50 & 80 & 100 & ImageNet-21K\\
\midrule
mIoU (\%)&
$61.8$&
$63.5$&
$64.2$&
$65.1$&
$65.6$&
$65.8$&
$66.6$\\
\bottomrule
\end{tabular}}
\end{table}

\begin{table*}[tb]
\centering
\caption{ \textbf{Detailed Architecture Specifications of PViT for Different Tasks}.
The same Standard Transformer ($12$ layers, $6$ heads, and $384$ hidden dimensions) is used as the backbone in all tasks. PViT uses a progressive tokenizer in segmentation tasks. For classification, since the number of input points is small, PViT only downsamples once.}
\label{tab:pvit_archi}%
\footnotesize
\resizebox{1.0\linewidth}{!}{\begin{tabular}{c|c|c|c}
\toprule
\textbf{Specification}&  \textbf{Classification}&    \textbf{Part Segmentation}&  \textbf{Semantic Segmentation}\\
\midrule
\textbf{Input Point Cloud}
&$1024\times3$
&$2048\times6$
&$16384\times6$\\
\midrule
\textbf{Tokenizer}& 
$\left[ {\begin{array}{c}
FPS (1/4)\\
KNN (32) \\
h_{1\mathbf\Theta}~MLP (384, 384)\\
h_{2\mathbf\Theta}~MLP (768, 384)\\
\end{array} } \right]
$
&
$\left[ {\begin{array}{c}
FPS (1/4)\\
KNN (32) \\
h_{11\mathbf\Theta}~MLP (192, 192)\\
h_{12\mathbf\Theta}~MLP (384, 192)\\
FPS (1/4)\\
KNN (32) \\
h_{21\mathbf\Theta}~MLP (384, 384)\\
h_{22\mathbf\Theta}~MLP (768, 384)\\
\end{array} } \right]
$
&
$\left[ {\begin{array}{c}
FPS (1/4)\\
KNN (32) \\
h_{11\mathbf\Theta}~MLP (192, 192)\\
h_{12\mathbf\Theta}~MLP (384, 192)\\
FPS (1/4)\\
KNN (32) \\
h_{21\mathbf\Theta}~MLP (384, 384)\\
h_{22\mathbf\Theta}~MLP (768, 384)\\
\end{array} } \right]
$
\\
\midrule
\textbf{Positional Embedding}
&
$MLP(128, 384)$
&
$MLP(128, 384)$
&
$MLP(128, 384)$
\\
\midrule
\textbf{Transformer Backbone}
&  
 (L=12, H=6, C=384)
& 
 (L=12, H=6, C=384)
&  
 (L=12, H=6, C=384)
\\
\midrule
\textbf{Decoder} &
$\left[ {\begin{array}{c}
concat(\text{MAX}; \text{CLS})\\
MLP(512, 256, 15)\\
\end{array} } \right]
$
&
$\left[ {\begin{array}{c}
Feature Propagation\\
Feature Propagation\\
concat(\mathbf{x}; \text{MAX}; \text{CLS})\\
MLP(256, 13)
\end{array} } \right]
$
&
$
\left[ {\begin{array}{c}
Feature Propagation\\
Feature Propagation\\
concat(\mathbf{x}; \text{MAX}; \text{CLS})\\
MLP(256, 50)
\end{array} } \right]
$
\\
\bottomrule
\end{tabular}}
\end{table*}%

\end{document}